\documentclass[final]{cvpr}

\usepackage[utf8]{inputenc} 
\usepackage[T1]{fontenc}    
\usepackage{mathptmx}
\usepackage{hyperref}       
\usepackage{url}            
\usepackage{booktabs}       
\usepackage{amsfonts}       
\usepackage{mathtools}
\usepackage{nicefrac}       
\usepackage{microtype}      
\usepackage{lipsum}
\usepackage{graphicx}
\usepackage{caption}
\usepackage{subcaption}
\usepackage{csquotes}
\usepackage{tablefootnote}
\usepackage[flushleft]{threeparttable}
\usepackage[english]{babel}
\usepackage{biblatex}
\graphicspath{ {./images/} }

\begin{document}

\title{TinaFace: Strong but Simple Baseline for Face Detection}

\author{
 Yanjia Zhu\thanks{Equal contribution.}, Hongxiang Cai\footnotemark[1], Shuhan Zhang\thanks{Data analysis.}, Chenhao Wang\footnotemark[2], Yichao Xiong\thanks{Corresponding author.} \\
  Media Intelligence Technology Co.,Ltd\\
  \texttt{\small \{yanjia.zhu, hongxiang.cai, shuhan.zhang, chenhao.wang, yichao.xiong\}@media-smart.cn} 
}

\maketitle
\begin{abstract}
Face detection has received intensive attention in recent years. Many works present lots of special methods for face detection from different perspectives like model architecture, data augmentation, label assignment and etc., which make the overall algorithm and system become more and more complex. In this paper, we point out that \textbf{there is no gap between face detection and generic object detection}. Then we provide a strong but simple baseline method to deal with face detection named TinaFace. We use ResNet-50 \cite{he2016deep} as backbone, and all modules and techniques in TinaFace are constructed on existing modules, easily implemented and based on generic object detection. On the hard test set of the most popular and challenging face detection benchmark WIDER FACE \cite{yang2016wider}, with single-model and single-scale, our TinaFace achieves 92.1\% average precision (AP), which exceeds most of the recent face detectors with larger backbone. And after using test time augmentation (TTA), our TinaFace outperforms the current state-of-the-art method and achieves 92.4\% AP. The code is available at \url{https://github.com/Media-Smart/vedadet/tree/main/configs/trainval/tinaface}.
\end{abstract}


\section{Introduction}

Face detection becomes a very important task in computer vision, since it is the first and fundamental step of most tasks and applications about faces, such as face recognition, verification, tracking, alignment, expression analysis etc.. Therefore, so many methods are presented in this field from different perspectives recently. Some works \cite{deng2019retinaface, earp2019face, yashunin2020maskface} introduce annotated landmarks information as extra supervision signal, and some of others \cite{zhang2020asfd, zhang2020refineface, tang2018pyramidbox, li2019pyramidbox++, Najibi_2017_ICCV, Najibi_2019_CVPR, zhang2017s3fd} pay more attention to the design of network. Besides, some new loss designs \cite{zhang2020asfd, zhang2020refineface, Li_2019_CVPR} and data augmentation methods \cite{li2019pyramidbox++, tang2018pyramidbox} are presented. What's more, a few works \cite{Liu_2020_CVPR, zhang2017s3fd} begin to redesign the matching strategy and label assignment process. Obviously, face detection seems to be gradually separated out from generic object detection and forms a new field.

\begin{figure*}[t]
\centering
  \includegraphics[width=1.0\linewidth]{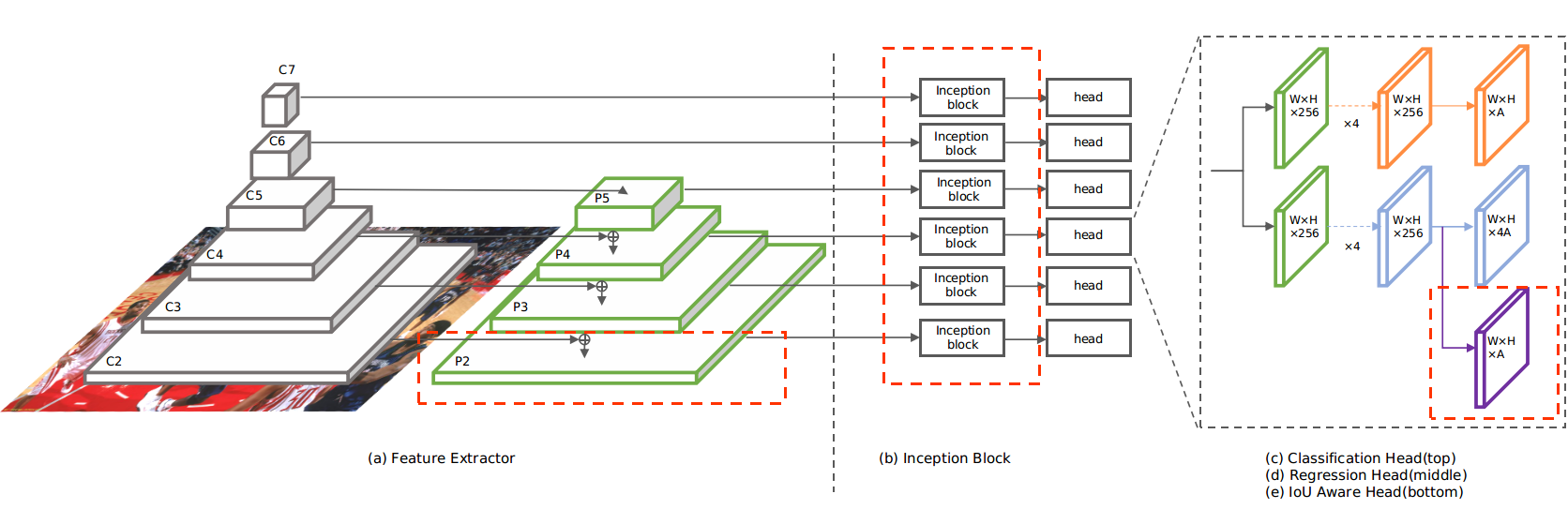}
\centering
\caption{The model architecture of TinaFace. (a) Feature Extractor: ResNet-50 \cite{he2016deep} and 6 level Feature Pyramid Network \cite{lin2017feature} to extract the multi-scale features of input image. (b) Inception block to enhance receptive field. (c) Classification Head: 5 layers FCN for classification of anchors. (d) Regression Head: 5 layers FCN for regression of anchors to ground-truth objects boxes. (e) IoU Aware Head: a single convolutional layer for IoU prediction.}
\label{fig1}
\end{figure*}

Intuitively, face detection is actually an application of generic object detection. To some degree, face is an object. So naturally there are a series of questions to be asked, \textit{"what is the difference between face detection and generic object detection?"}, \textit{"Why not using generic object detection techniques to deal with face detection?"}, and \textit{"is it necessary to additionally design special methods for handling face detection?"}.

First, from the perspective of data, the properties that faces own also exist in objects, like pose, scale, occlusion, illumination, blur and etc.. And the unique properties in faces like expression and makeup can also correspond to distortion and color in objects. Then from the perspective of challenges encountered by face detection like multi-scale, small faces and dense scenes, they all exist in generic object detection. Thus, face detection seems to be just a subproblem of generic object detection. To better and further answer above questions, we provide a simple baseline method based on generic object detection to outperform the current state-of-the-art methods on the hard test set of WIDER FACE \cite{yang2016wider}.

The main contributions of this work can be summarized as:
\begin{itemize}
\item Indicating that face detection is actually a one class generic object detection problem and can be handled by techniques in generic object detection.
\item Providing a strong but simple baseline method for face detection named TinaFace. All ideas and modules used in TinaFace are based on generic object detection.
\item With single-scale and single-model, we achieve 92.1\% average precision(AP) in hard settings on the test subset of WIDER FACE, which already exceed most of recent methods with larger backbone and Test Time Augmentation (TTA). Our final model gets 92.4\% AP in hard settings on the test subset and outperforms current state-of-the-art methods for face detection.
\end{itemize}

\section{Related Work}

\paragraph{Generic Object Detection.}
Generic object detection aims at locating and classifying the existing objects in the given picture. Before the booming of deep learning, generic object detection is mainly based on the hand-crafted feature descriptors like SIFT \cite{lowe2004distinctive} and HOG \cite{dalal2005histograms}. And the most successful methods like DPM \cite{felzenszwalb2008discriminatively} combine multi-scale hand-crafted features, sliding window, deformable part and SVM classifier to form a generic object detector.

With AlexNet \cite{krizhevsky2017imagenet} winning the championship of Large Scale Visual Recognition Challenge 2012 (ILSVRC2012) by a large gap, the era of deep learning is coming, and generic object detection has been quickly dominated by deep learning methods. Two-stage methods start from R-CNN \cite{girshick2014rich} and Fast R-CNN \cite{girshick2015fast}. And soon Faster R-CNN \cite{ren2015faster} proposes RPN network to replace the selective search to generate proposals by pre-define anchors, which becomes the most classical anchor-based generic object detection method. Based on Faster R-CNN \cite{ren2015faster}, there are so many new methods presented like FPN \cite{lin2017feature}, Mask R-CNN \cite{he2017mask}, Cascade R-CNN \cite{cai2018cascade} and etc.. In order to overcome the high latency of two-stage methods, many one-stage methods are presented like series of YOLO \cite{redmon2016you, redmon2017yolo9000, redmon2018yolov3}, SSD \cite{liu2016ssd} and RetinaNet \cite{lin2017focal}. To handling the multiple scale or small objects problem, YOLOs \cite{redmon2016you, redmon2017yolo9000, redmon2018yolov3} present novel anchor matching strategy including consideration of feedback of proposals and one ground-truth vs. one anchor, and also reweight the regression of width and height of objects. Then SSD \cite{liu2016ssd} uses a hierarchy of backbone features, while FPN \cite{lin2017feature} presents feature pyramids. Besides, the series of SNIP \cite{singh2018analysis} and SNIPER \cite{singh2018sniper}, multi-scale training and multi-scale testing can also deal with the multiple scale problem. 

In addition to the new method proposed in generic object detection, developments in other fields, like normalization methods and deep convolutional networks, also promote generic object detection. Batch normalization (BN) \cite{ioffe2015batch} normalizes features within a batch along channel dimension, which can help models converge and enable models to train. In order to handle the dependency with batch size of BN, group normalization (GN) \cite{wu2018group} divides the channels into groups and computes within each group the mean and variance for normalization. Then for deep convolutional networks, after AlexNet \cite{krizhevsky2017imagenet}, VGG \cite{simonyan2014very} increases depth using an architecture with very small $3\times3$ convolution filters, GoogLeNet \cite{szegedy2015going} introduces Inception modules to use different numbers of small filters in parallel to form features of different receptive fields and help model to capture objects as well as context at multiple scales, and ResNet \cite{he2016deep} demonstrates the importance of the original information flow and presents skip connection to handle the degradation with deeper networks.

\paragraph{Face Detection.}
As an application of generic object detection, the history of face detection is almost the same. Before the era of deep learning, face detectors are also based on hand-crafted features like Haar \cite{viola2001robust}. After the most popular and challenging face detection benchmark WIDER FACE dataset \cite{yang2016wider} presented, face detection develops rapidly focusing on the extreme and real variation problem including scale, pose, occlusion, expression, makeup, illumination, blur and etc.. Almost all the recent face detection methods evolve from the existing generic object detection methods. Based on SSD \cite{liu2016ssd}, S$^3$FD \cite{zhang2017s3fd} extends anchor-associated layers to C3 stage and proposes a scale compensation anchor matching strategy in order to cover the small faces, PyramidBox \cite{tang2018pyramidbox} proposes PyramidAnchors (PA), Low-level Feature Pyramid Networks (LFPN), Context-sensitive Predict Module (CPM) to emphasize the importance of context and data-anchor-sampling augmentation to increase smaller faces, and DSFD \cite{Li_2019_CVPR} introduce a dual-shot detector using Improved Anchor Matching (IAM) and Progressive Anchor Loss (PAL). Then Based on RetinaNet \cite{lin2017focal}, RetinaFace \cite{deng2019retinaface} manually annotates five facial landmarks on faces to serve as extra supervision signal, RefineFace \cite{zhang2020refineface} introduces five extra modules Selective Two-step Regression (STR), Selective Two-step Classification (STC), Scale-aware Margin Loss (SML), Feature Supervision Module (FSM) and Receptive Field Enhancement (RFE), and HAMBox \cite{Liu_2020_CVPR} emphasize the strong regression ability of some unmatched anchors and present an Online High-quality Anchor Mining Strategy (HAMBox). Besides, ASFD \cite{zhang2020asfd} uses neural architecture search technique to automatically search the architecture for efficient multi-scale feature fusion and context enhancement.

To sum up, methods presented in face detection almost cover every part of deep learning training from data processing to loss designs. It is obvious that all of these methods focus on the challenge of small faces. However, actually there are so many methods in generic object detection, which we mention above, solving this problem. Therefore, based on some of these methods, we present TinaFace, a strong but simple baseline method for face detection.

\section{TinaFace}
\label{sec3}
Basically, we start from the one-stage detector RetinaNet \cite{lin2017focal} as some previous works do. The architecture of TinaFace is shown in Figure \ref{fig1} where the red dashed boxes demonstrate the different parts from RetinaNet \cite{lin2017focal}.

\subsection{Deformable Convolution Networks}
\label{sec3:sub1}
There is an inherent limitation in convolution operation, that is, we feed it with a strong prior about the sampling position which is fixed and rigid. Therefore, it is hard for networks to learn or encode complex geometric transformations, and the capability of models is limited. In order to further improve the capability of our model, we employ DCN \cite{dai2017deformable} into the stage four and five of the backbone.

\subsection{Inception Module}
\label{sec3:sub2}
Multi-scale is always a challenge in generic object detection. The most common ways to deal with it are multi-scale training, FPN architecture and multi-scale testing. Besides, we employ inception module \cite{szegedy2015going} in our model to further enhance this ability. The inception module uses different numbers of $3\times3$ convolutional layers in parallel to form features of different receptive fields and then combine them, which help model to capture objects as well as context at multiple scales.

\subsection{IoU-aware Branch}
\label{sec3:sub3}
IoU-aware \cite{wu2020iou} is an extremely simple and elegant method to relieve the mismatch problem between classification score and localization accuracy of a single-stage object detector, which can help resort the classification score and suppress the false positive detected boxes (high score but low IoU). The architecture of IoU-aware is shown in Figure \ref{fig1}, and the only difference is the purple part, a parallel head with a regression head to predict the IoU between the detected box and the corresponding ground-truth object. And this head only consists of a single $3 \times 3$ convolution layer, followed by a sigmoid activation layer. At the inference phase, the final detection confidence is computed by following equation,
\begin{equation}
score = p_i^\alpha IoU_i^{(1-\alpha)} 
\end{equation}
where $p_i$ and $IoU_i$ are the original classification score and predicted IoU of $i$th detected box, and $\alpha \in [0, 1]$ is the hyperparameter to control the contribution of the classification score and predicted IoU to the final detection confidence. 

\subsection{Distance-IoU Loss}
\label{sec3:sub4}
The most common loss used in bbox regression is Smooth L1 Loss \cite{girshick2015fast} , which regresses the parameterizations of the four coordinates (box's center and its width and height). However, these optimization targets are not consistent with the regression evaluation metric IoU, that is, lower loss is not equivalent with higher IoU. Therefore, we turn to different IoU losses presented in past few years, directly regressing the IoU metric, such as GIoU \cite{rezatofighi2019generalized}, DIoU and CIoU \cite{zheng2020distance}. The reason we choose DIoU \cite{zheng2020distance} as our regression loss is that  small faces is the main challenge of face detection since there are about two thirds data in WIDER FACE \cite{yang2016wider} belong to small object and DIoU \cite{zheng2020distance} is more friendly to small objects. Practically, DIoU gets better performance on APsmall of the validation set of MS COCO 2017 \cite{lin2014microsoft}. And theoretically, DIoU is defined as: 
\begin{equation}
L_{DIoU} = 1 - IoU + \frac{\rho^2(\textbf{b}, \textbf{b}^{gt})}{c^2}
\end{equation}
where $\textbf{b}$ and $\textbf{b}^{gt}$ denote the central points of predicted box and ground-truth box, $\rho(\cdot)$ is the Euclidean distance, and c is the diagonal length of the smallest enclosing box covering the two boxes. The extra penalty term $\frac{\rho^2(\textbf{b}, \textbf{b}^{gt})}{c^2}$ proposes to minimize the normalized distance between central points of predicted box and ground-truth box. Compared to large objects, the same distance of central points in small objects will be penalized more, which help detectors learn more about small objects in regression.

\begin{figure*}[ht]
\centering                                                                               
\begin{subfigure}{.5\textwidth}
  \centering
  \includegraphics[width=\linewidth]{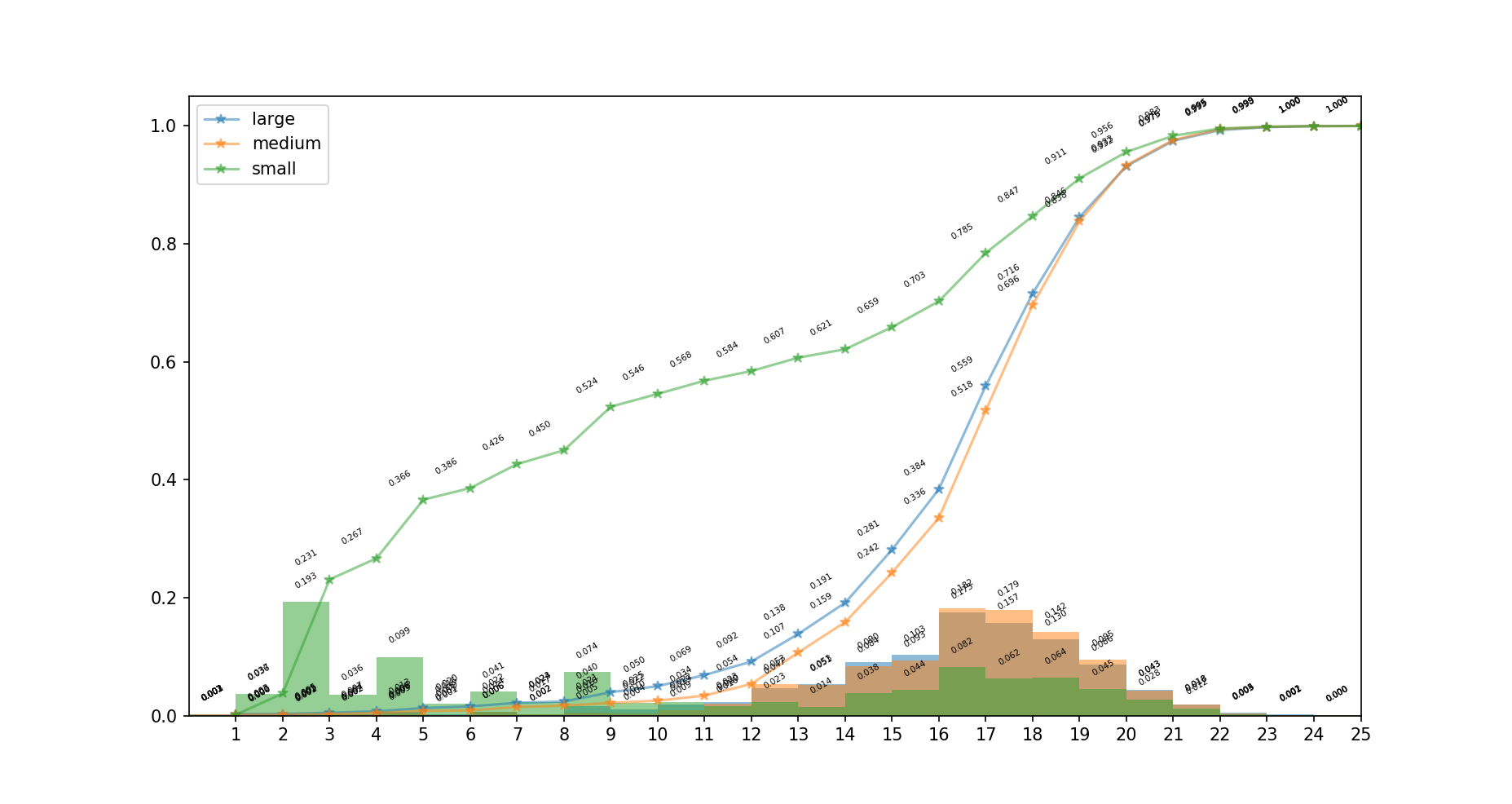}
  \caption{}
  \label{fig2:sub1}
\end{subfigure}%
\begin{subfigure}{.5\textwidth}
  \centering
  \includegraphics[width=\linewidth]{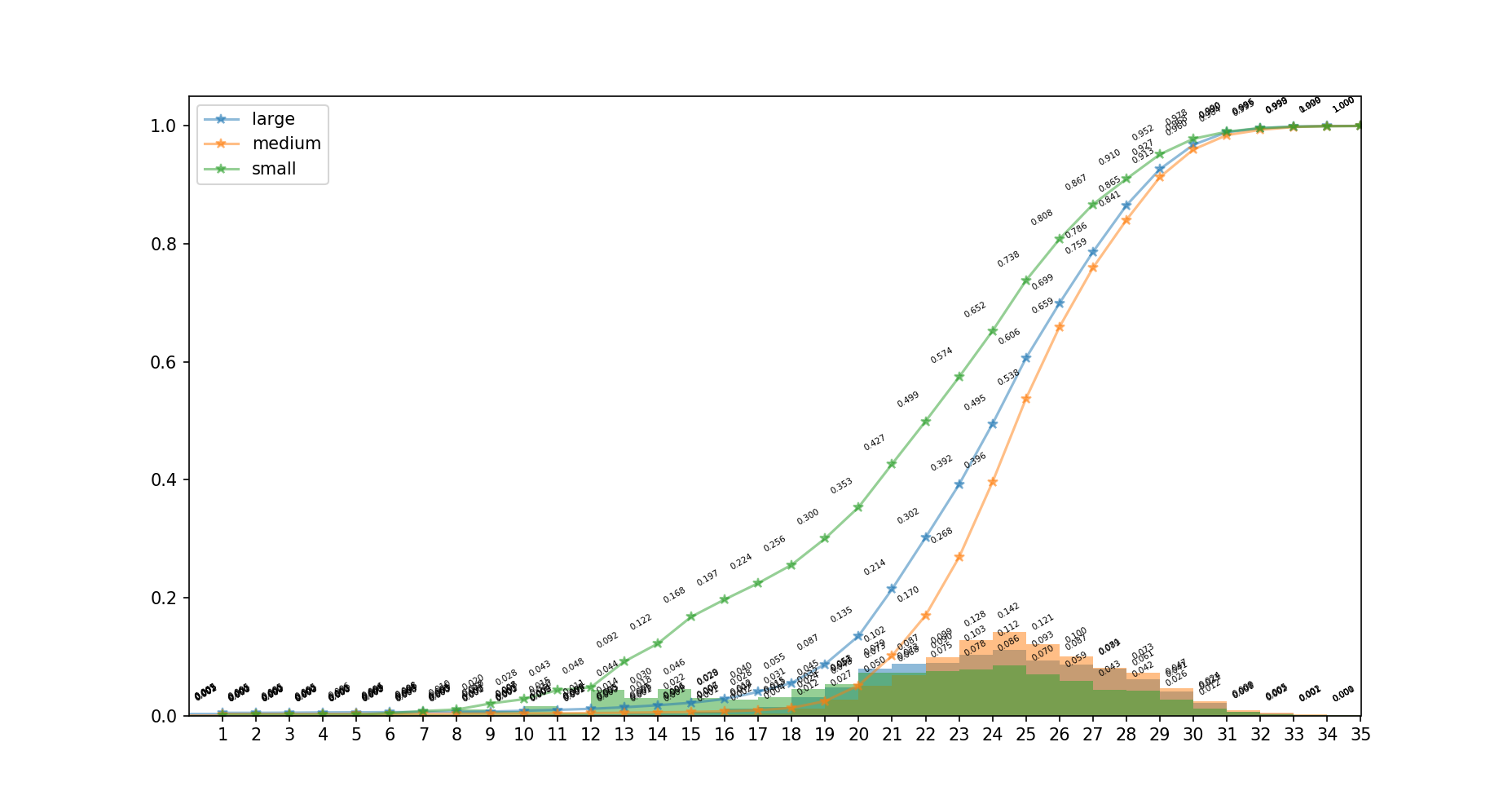}
  \caption{}
  \label{fig2:sub2}
\end{subfigure}
\caption{The cumulative distribution and density function of the number of positive samples assigned to each ground-truth. Different colors represent different scales of ground-truth based on the evaluation across scales on COCO dataset. (a) distribution of Retinaface's \cite{deng2019retinaface} settings. (b) distribution of this work's settings. }
\label{fig:fig2}
\end{figure*}

\begin{table*}[ht]
 \caption{AP performance on WIDER FACE validation subset}
  \centering
  \begin{tabular}{ c c c c c c | c c c }
    \toprule
    Baseline & DIoU & Inception &  IoU-aware & DCN & TTA & Easy & Medium & Hard \\
    \midrule
    $\surd$ & - & - & - & - & - & 0.959 & 0.952 & 0.924  \\
    $\surd$ & $\surd$ & - & - & - & - & 0.959 & 0.952 & 0.927  \\
    $\surd$ & $\surd$ & $\surd$ & - & - & - & 0.958 & 0.952 & 0.928  \\
    $\surd$ & $\surd$ & $\surd$ & $\surd$ & - & - & 0.963 & 0.955 & 0.929  \\
    $\surd$ & $\surd$ & $\surd$ & $\surd$ & $\surd$ & - & 0.963 & 0.957 & 0.930  \\
    $\surd$ & $\surd$ & $\surd$ & $\surd$ & $\surd$ & $\surd$ & \textbf{0.970} & \textbf{0.963} & \textbf{0.934}  \\
    \bottomrule
  \end{tabular}
  \label{tab:table1}
\end{table*}

\begin{table*}[ht]
 \caption{AP performance of different methods on WIDER FACE validation subset and test subset}
  \centering
    \begin{threeparttable}
      \begin{tabular}{ c c c c c c c c }
        \toprule
         &  &  & Val &  &  & Test & \\
        Method & Backbone & Easy & Medium & Hard & Easy & Medium & Hard \\
        \cmidrule(r){1-1} \cmidrule(r){2-2} \cmidrule(r){3-5} \cmidrule(r){6-8}
        AInnoFace \cite{zhang2019accurate} \textdagger & ResNet-152 & 0.970 & 0.961 & 0.918 & 0.965 & 0.957 & 0.912  \\
        RetinaFace \cite{deng2019retinaface} \textdagger & ResNet-152 & 0.969 & 0.961 & 0.918 & 0.963 & 0.956 & 0.914  \\
        RefineFace \cite{zhang2020refineface} \textdagger & ResNet-152 & 0.972 & 0.962 & 0.920 & 0.966 & 0.958 & 0.914  \\
        ASFD-D6 \cite{zhang2020asfd} \textdagger & ResNet-152 & \textbf{0.972} & \textbf{0.965} & 0.925 & \textbf{0.967} & \textbf{0.962} & 0.921  \\
        HAMBox \cite{Liu_2020_CVPR} \textdagger & ResNet-50 & 0.970 & 0.964 & 0.933 & 0.959 & 0.955 & 0.923 \\
        \cmidrule(r){1-1} \cmidrule(r){2-2} \cmidrule(r){3-5} \cmidrule(r){6-8}
        TinaFace \textbf{(ours)} & ResNet-50 & 0.963 & 0.957 & 0.930 & 0.952 & 0.947 & 0.921 \\
        TinaFace \textbf{(ours)} \textdagger & ResNet-50 & 0.970 & 0.963 & \textbf{0.934} & 0.958 & 0.953 & \textbf{0.924} \\
        \bottomrule
      \end{tabular}
      \begin{tablenotes}
        \item[\textdagger] Note that different methods may use different TTA methods. (It is difficult to verify since most of these methods do not provide codes).
      \end{tablenotes}
    \end{threeparttable}
  \label{tab:table2}
\end{table*}

\section{Experiments}

\subsection{Dataset}
WIDER FACE dataset \cite{yang2016wider} is the largest face detection dataset, which contains 32,203 images and 393,703 faces. Since its variety of scale, pose, occlusion, expression, illumination and event, it is difficult and close to reality. The whole dataset is divided into train/val/test sets by ratio 50\%/10\%/40\% within each event class. Furthermore, based on the detection rate of EdgeBox \cite{zitnick2014edge}, each subset is defined into three levels of difficulty: ’Easy’, ’Medium’, ’Hard’. From the name of these three levels, we know that 'Hard' is more challenging. And from further analysis, we find that data in 'Hard' covers 'Medium' and 'Easy', which demonstrate that performance on 'Hard' can better reflect the effectiveness of different methods.

\subsection{Implementation Details}

\paragraph{Feature Extractor.}
We use ResNet-50 \cite{he2016deep} as backbone and Feature Pyramid Network (FPN) \cite{lin2017feature} as neck to construct the feature extractor. This combination is widely used in almost all detectors, so we think it can serve as a fair playground for replication and comparison. In order to cover the tiny faces, FPN \cite{lin2017feature} we employed extends to level $P_2$ like some previous works do. In total, there are 6 levels in FPN \cite{lin2017feature} from level $P_2$ to $P_7$.

\paragraph{Losses.}
The losses of classification, regression and IoU prediction are focal loss, DIoU loss and cross-entropy loss, respectively. 

\paragraph{Normalization Method}
Batch Normalization (BN) \cite{ioffe2015batch} is an extremely important technique for deep learning. It can help models converge and enable various networks to train. However, the performance of the model will degrade with the batch size decreasing especially when batch size is smaller than 4, caused by inaccurate batch statistics estimation. Considering that large volume GPUs are not widely used, which may cause problems for replication, with GeForce GTX 1080 Ti, we replace all the BN layer in network with Group Normalization \cite{wu2018group} which is a simple alternative to BN and independent of batch sizes, and the performance of which is stable.

\paragraph{Anchor and Assigner Settings}
Basically, we set 6 anchors from the set $2^{4/3} \times \{4, 8, 16, 32, 64, 128\}$ since there are 6 levels in our FPN \cite{lin2017feature}. We adjust the base scale to $2^{4/3}$ in order to better cover the tiny faces, use the mean value of aspect ratio of ground-truths as anchor ratio, and set three scales at step $2^{1/3}$ in each level. For assigner, the IoU threshold for matching strategy is 0.35, and ignore-zone is not applied.

To better understand the advantage of our settings, we utilize the detection analysis tool \footnote{\url{https://github.com/Media-Smart/volkscv}} and conduct two experiments to get the distribution of positive samples assigned to each ground-truth shown in Figure \ref{fig:fig2}. As illustrated in Figure \ref{fig2:sub1}, although RetinaFace \cite{deng2019retinaface} can recall most of the faces, it does not pay attention to the imbalance problem across scales, that is, small ground-truths get less positive anchors to train, while large one can get more, which leads the degraded performance on small ground-truths. Turning to Figure \ref{fig2:sub2}, we notice that the imbalanced problem is largely relieved. The distribution of the number of positive assigned samples is highly similar across scale.

\paragraph{Data Augmentation.}
First, crop the square patch from the original picture with a random size from the set $[0.3, 0.45, 0.6, 0.8, 1.0]$ of the short edge of the original image and keep the overlapped part of the face box if its centre is within the crop patch. Then do photo distortion and random horizontal flip with the probability of 0.5. Finally, resize the patch into $640 \times 640$ and normalize.

\paragraph{Training Settings.}
We train the model by using SGD optimizer (momentum 0.9, weight decay 5e-4) with batch size $3\times 4$ on three GeForce GTX 1080 Ti. The schedule of learning rate is annealing down from 3.75e-3 to 3.75e-5 every 30 epochs out of 630 epochs using the cosine decay rule. And in the first 500 iterations, learning rate linearly warms up from 3.75e-4 to 3.75e-3.

\paragraph{Testing Settings.}
Single Scale testing only contains a keep-ratio resize, which guarantees that the short and long edge of image do not surpass 1100 and 1650. Test Time Augmentation(TTA) is composed of multi-scale (the short edge of image is $[500, 800, 1100, 1400, 1700]$), shift (the direction is $[(0, 0), (0, 1), (1, 0), (1, 1)]$), horizontal flip and box voting.

\subsection{Evaluation on WIDER FACE}
As shown in Table \ref{tab:table1}, we present the AP performance of models described in Section \ref{sec3} on WIDER FACE validation subset. Our baseline model using single scale testing gets 95.9\%, 95.2\%, 92.4\% in the three settings on the validation subset. Then we introduce DIoU \cite{zheng2020distance}, Inception \cite{szegedy2015going}, IoU-aware \cite{wu2020iou}, DCN \cite{dai2017deformable} modules and TTA to further improve the performance of detector by 1.1\%, 1.1\%, 1.0\% on three settings, respectively.

\begin{figure*}[ht]
\centering
\begin{subfigure}{.5\textwidth}
  \centering
  \includegraphics[width=\linewidth]{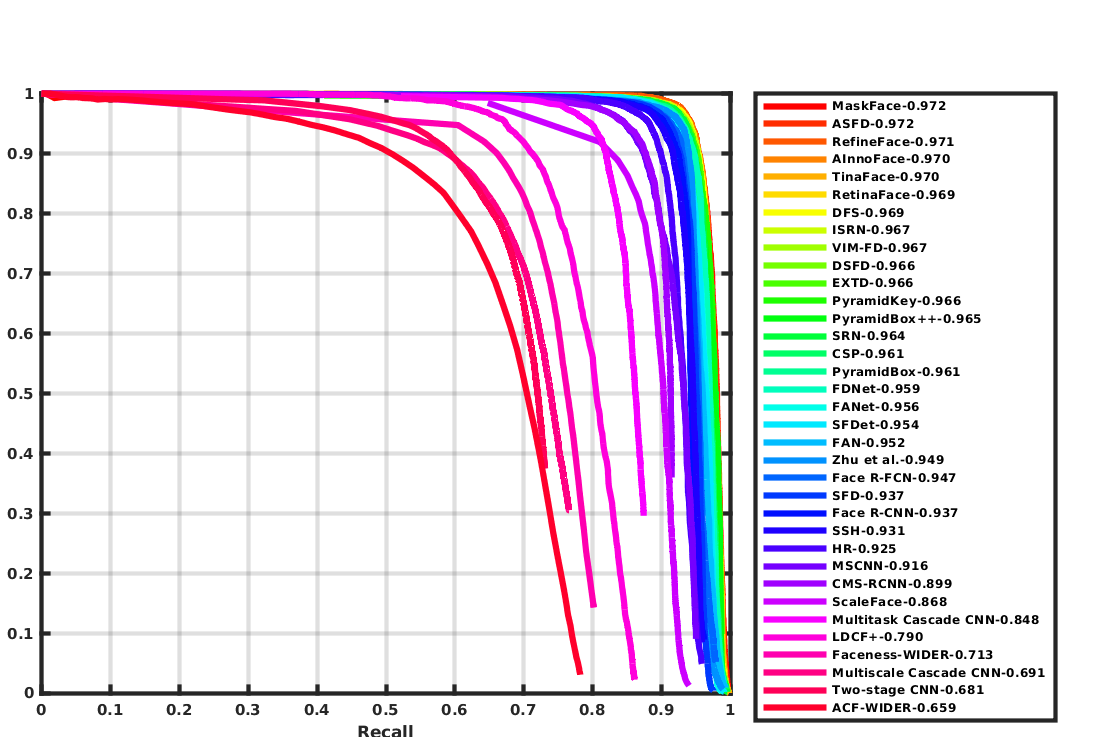}
  \caption{Val: Easy}
  \label{fig3:sub1}
\end{subfigure}%
\begin{subfigure}{.5\textwidth}
  \centering
  \includegraphics[width=\linewidth]{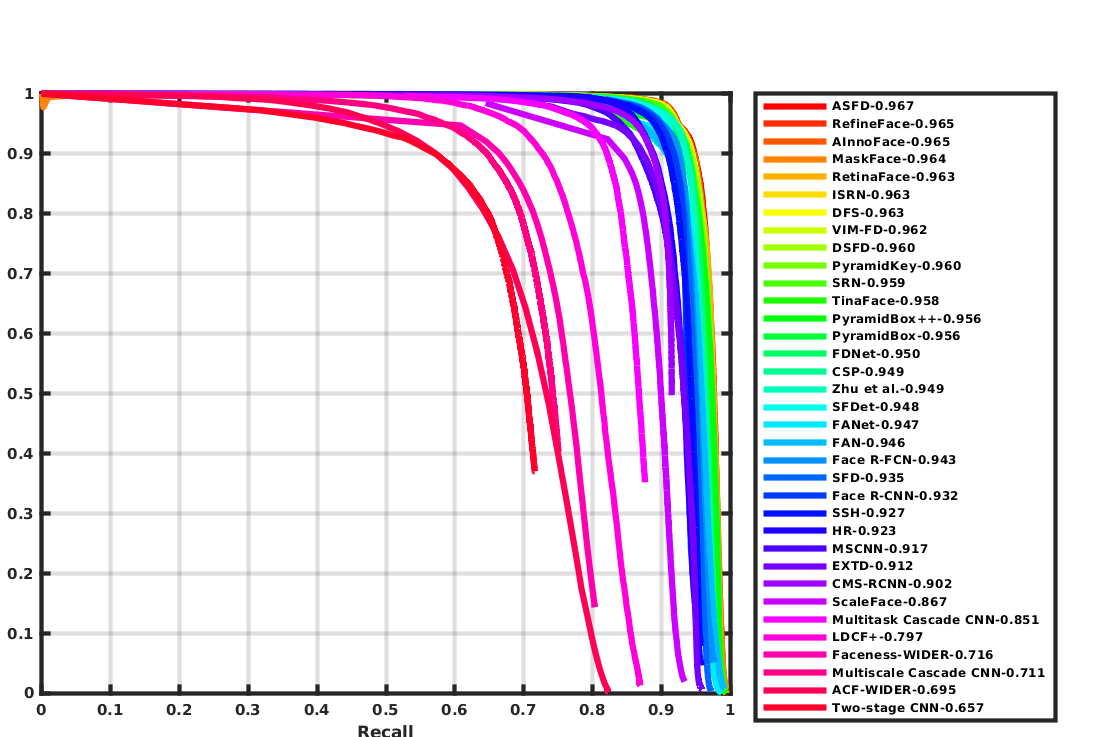}
  \caption{Test: Easy}
  \label{fig3:sub2}
\end{subfigure}

\begin{subfigure}{.5\textwidth}
  \centering
  \includegraphics[width=\linewidth]{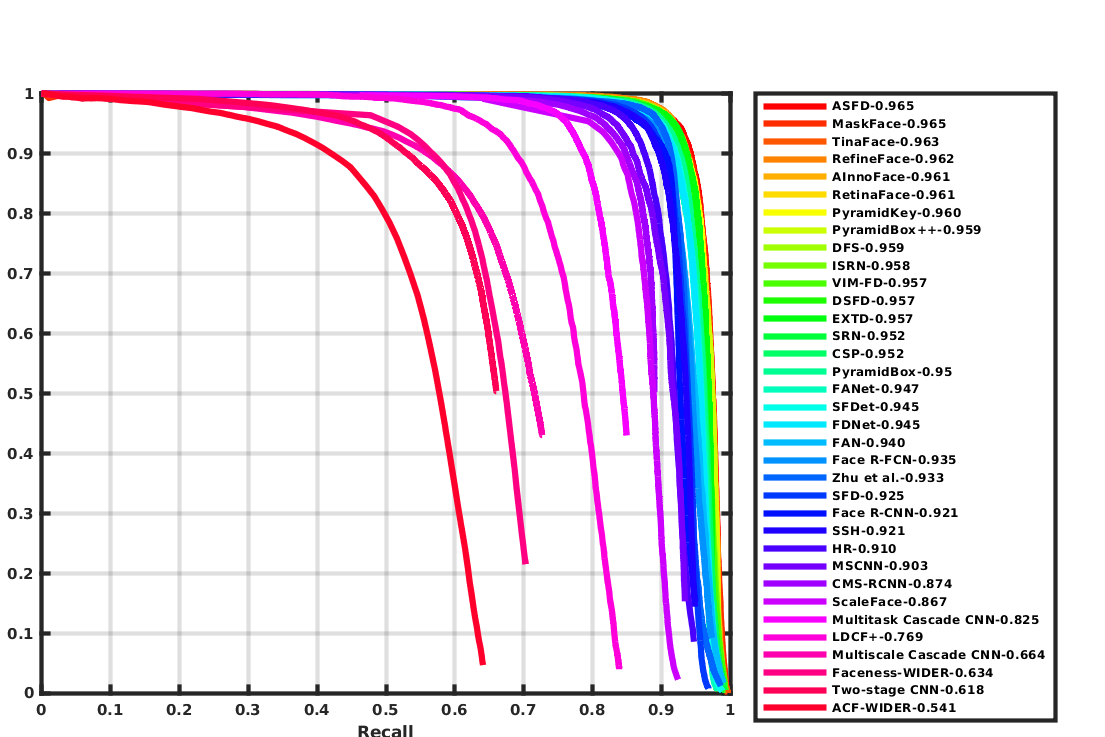}
  \caption{Val: Medium}
  \label{fig3:sub3}
\end{subfigure}%
\begin{subfigure}{.5\textwidth}
  \centering
  \includegraphics[width=\linewidth]{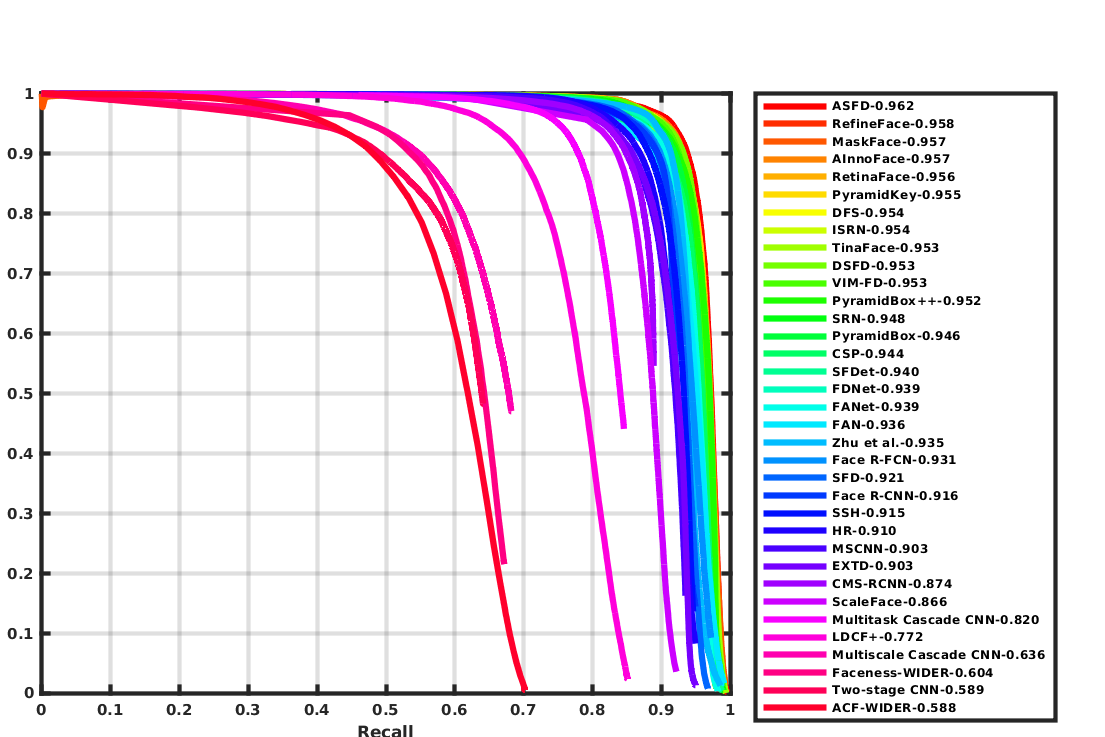}
  \caption{Test: Medium}
  \label{fig3:sub4}
\end{subfigure}

\begin{subfigure}{.5\textwidth}
  \centering
  \includegraphics[width=\linewidth]{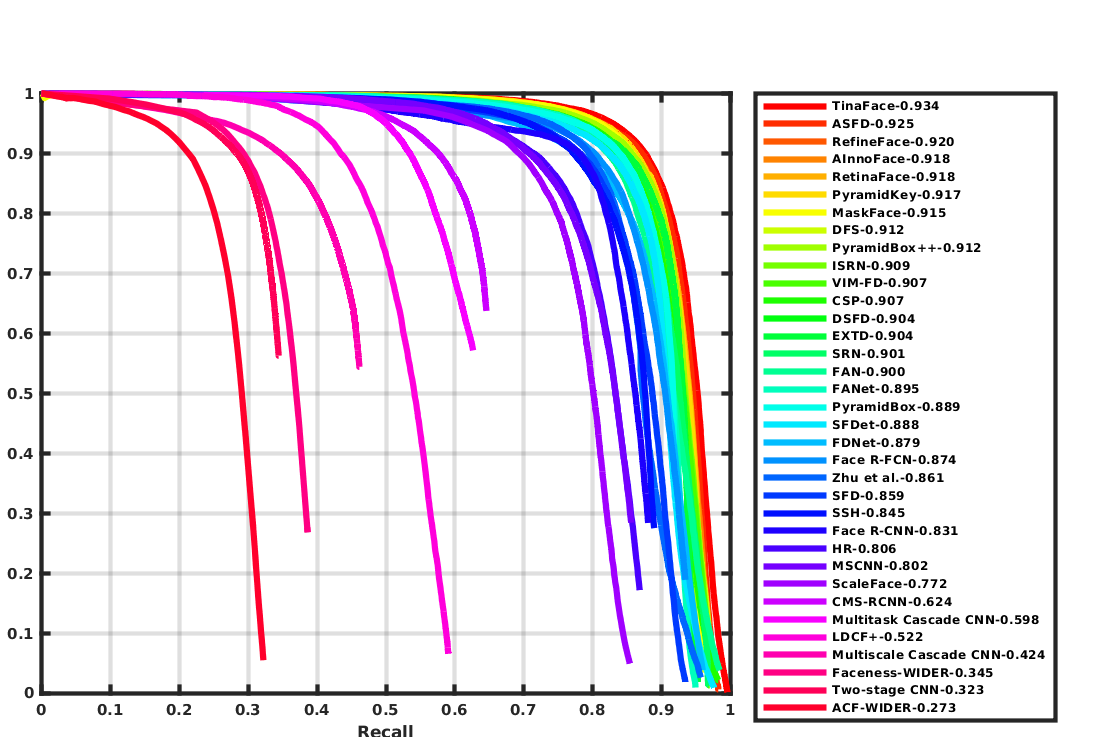}
  \caption{Val: Hard}
  \label{fig3:sub5}
\end{subfigure}%
\begin{subfigure}{.5\textwidth}
  \centering
  \includegraphics[width=\linewidth]{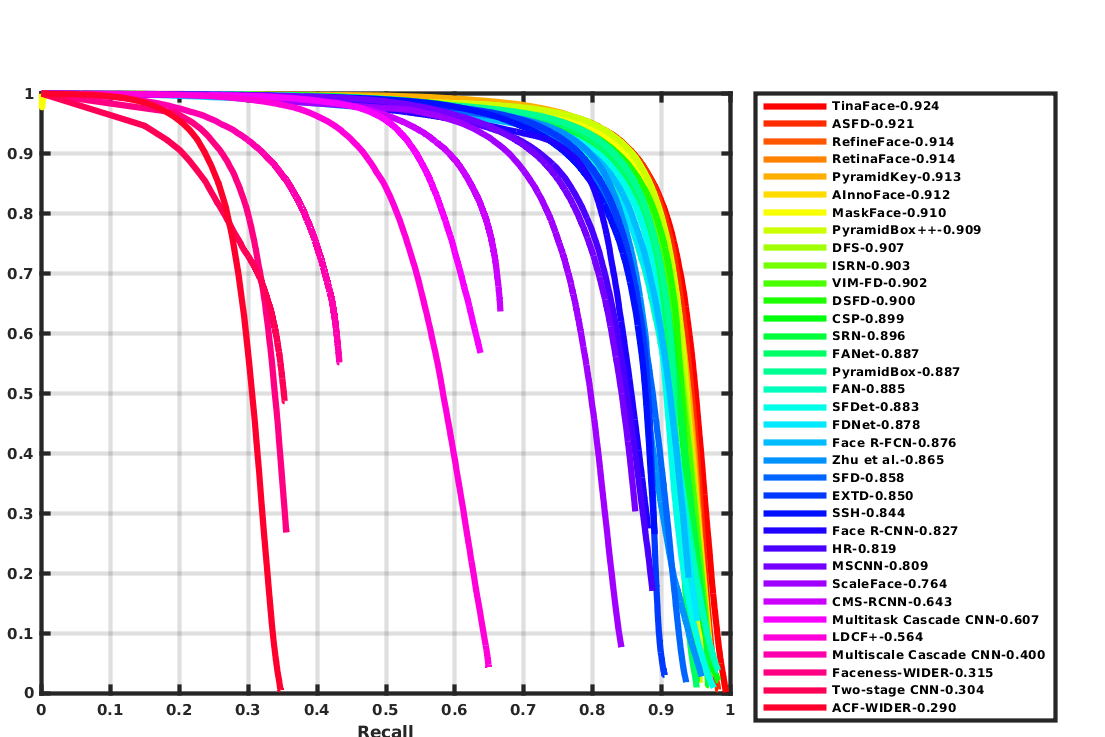}
  \caption{Test: Hard}
  \label{fig3:sub6}
\end{subfigure}

\caption{Precision-recall curves on the WIDER FACE validation and test subsets.}
\label{fig:fig3}
\end{figure*}

\subsection{Comparsion with other methods on WIDER FACE}
As shown in Figure \ref{fig:fig3}, we compare TinaFace with recent face detection methods \cite{zhang2020asfd, zhang2020refineface, zhang2019accurate, deng2019retinaface, earp2019face, yashunin2020maskface, tian2018learning, li2019pyramidbox++, zhang2019improved, zhang2019robust, Liu_2019_CVPR, Li_2019_CVPR, yoo2019extd, chi2019selective, wang2017face, zhang2020feature, tang2018pyramidbox, zhang2019single, zhang2018face, wang2017detecting, Zhu_2018_CVPR, zhang2017s3fd, Najibi_2017_ICCV, wang2017facercnn, Hu_2017_CVPR, cai2016unified, yang2017face, zhu2017cms, zhang2016joint, ohn2016boost, yang2016wider, yang2015facial, yang2014aggregate} on both validation and testing subsets. For better comparsion, we pick up top-5 methods to form the Table \ref{tab:table2} (HAMBox \cite{Liu_2020_CVPR} isn't listed in Figure \ref{fig:fig3} since its results are not updated on the official website of WIDER FACE \footnote{\url{http://shuoyang1213.me/WIDERFACE/WiderFace_Results.html}}).

Surprisingly, with single-scale and single-model, our model already gets very promising and almost state-of-the-art performance especially in the hard setting, which respectively outperforms ASFD-D6 \cite{zhang2020asfd} in validation subset and test subset. Moreover, our model uses ResNet-50 as backbone, which is much smaller than what ASFD-D6 \cite{zhang2020asfd} uses. In the case of using the same backbone, our final model with TTA outperforms the current state-of-the-art method HAMBox \cite{Liu_2020_CVPR}.

\section{Conclusion}
In this paper, we point out that face detection is actually a one class generic object detection problem. It indicates that methods presented in generic object detection can be used for handling this problem. Then we present a strong but simple baseline method based on generic object detection for dealing with face detection named TinaFace to further  illustrate this point. The whole network is simple and straightforward, and all the recent tricks equipped are easily implemented and built on existing modules. On the hard setting of the test subset of WIDER FACE, Our model without TTA already exceeds most recent face detection methods like ASFD-D6, which will be extremely efficient and effective. Besides, our final model achieves the state-of-the-art face detection performance.

\printbibliography

@inproceedings{lin2017focal,
  title={Focal loss for dense object detection},
  author={Lin, Tsung-Yi and Goyal, Priya and Girshick, Ross and He, Kaiming and Doll{\'a}r, Piotr},
  booktitle={Proceedings of the IEEE international conference on computer vision},
  pages={2980--2988},
  year={2017}
}

@inproceedings{yang2016wider,
  title={Wider face: A face detection benchmark},
  author={Yang, Shuo and Luo, Ping and Loy, Chen-Change and Tang, Xiaoou},
  booktitle={Proceedings of the IEEE conference on computer vision and pattern recognition},
  pages={5525--5533},
  year={2016}
}

@article{deng2019retinaface,
  title={Retinaface: Single-stage dense face localisation in the wild},
  author={Deng, Jiankang and Guo, Jia and Zhou, Yuxiang and Yu, Jinke and Kotsia, Irene and Zafeiriou, Stefanos},
  journal={arXiv preprint arXiv:1905.00641},
  year={2019}
}

@article{earp2019face,
  title={Face Detection with Feature Pyramids and Landmarks},
  author={Earp, Samuel WF and Noinongyao, Pavit and Cairns, Justin A and Ganguly, Ankush},
  journal={arXiv preprint arXiv:1912.00596},
  year={2019}
}

@article{yashunin2020maskface,
  title={MaskFace: multi-task face and landmark detector},
  author={Yashunin, Dmitry and Baydasov, Tamir and Vlasov, Roman},
  journal={arXiv preprint arXiv:2005.09412},
  year={2020}
}

@article{zhang2020asfd,
  title={ASFD: Automatic and Scalable Face Detector},
  author={Zhang, Bin and Li, Jian and Wang, Yabiao and Tai, Ying and Wang, Chengjie and Li, Jilin and Huang, Feiyue and Xia, Yili and Pei, Wenjiang and Ji, Rongrong},
  journal={arXiv preprint arXiv:2003.11228},
  year={2020}
}

@article{zhang2020refineface,
  title={Refineface: Refinement neural network for high performance face detection},
  author={Zhang, Shifeng and Chi, Cheng and Lei, Zhen and Li, Stan Z},
  journal={IEEE Transactions on Pattern Analysis and Machine Intelligence},
  year={2020},
  publisher={IEEE}
}

@inproceedings{tang2018pyramidbox,
  title={Pyramidbox: A context-assisted single shot face detector},
  author={Tang, Xu and Du, Daniel K and He, Zeqiang and Liu, Jingtuo},
  booktitle={Proceedings of the European Conference on Computer Vision (ECCV)},
  pages={797--813},
  year={2018}
}

@article{li2019pyramidbox++,
  title={Pyramidbox++: High performance detector for finding tiny face},
  author={Li, Zhihang and Tang, Xu and Han, Junyu and Liu, Jingtuo and He, Ran},
  journal={arXiv preprint arXiv:1904.00386},
  year={2019}
}

@InProceedings{Najibi_2019_CVPR,
author = {Najibi, Mahyar and Singh, Bharat and Davis, Larry S.},
title = {FA-RPN: Floating Region Proposals for Face Detection},
booktitle = {Proceedings of the IEEE/CVF Conference on Computer Vision and Pattern Recognition (CVPR)},
month = {June},
year = {2019}
}

@inproceedings{zhang2017s3fd,
  title={S3fd: Single shot scale-invariant face detector},
  author={Zhang, Shifeng and Zhu, Xiangyu and Lei, Zhen and Shi, Hailin and Wang, Xiaobo and Li, Stan Z},
  booktitle={Proceedings of the IEEE international conference on computer vision},
  pages={192--201},
  year={2017}
}

@InProceedings{Li_2019_CVPR,
author = {Li, Jian and Wang, Yabiao and Wang, Changan and Tai, Ying and Qian, Jianjun and Yang, Jian and Wang, Chengjie and Li, Jilin and Huang, Feiyue},
title = {DSFD: Dual Shot Face Detector},
booktitle = {Proceedings of the IEEE/CVF Conference on Computer Vision and Pattern Recognition (CVPR)},
month = {June},
year = {2019}
}

@InProceedings{Liu_2020_CVPR,
author = {Liu, Yang and Tang, Xu and Han, Junyu and Liu, Jingtuo and Rui, Dinger and Wu, Xiang},
title = {HAMBox: Delving Into Mining High-Quality Anchors on Face Detection},
booktitle = {IEEE/CVF Conference on Computer Vision and Pattern Recognition (CVPR)},
month = {June},
year = {2020}
}

@inproceedings{ioffe2015batch,
  title={Batch Normalization: Accelerating Deep Network Training by Reducing Internal Covariate Shift},
  author={Ioffe, Sergey and Szegedy, Christian},
  booktitle={International Conference on Machine Learning},
  pages={448--456},
  year={2015}
}

@inproceedings{wu2018group,
  title={Group normalization},
  author={Wu, Yuxin and He, Kaiming},
  booktitle={Proceedings of the European conference on computer vision (ECCV)},
  pages={3--19},
  year={2018}
}

@inproceedings{zheng2020distance,
  title={Distance-IoU Loss: Faster and Better Learning for Bounding Box Regression.},
  author={Zheng, Zhaohui and Wang, Ping and Liu, Wei and Li, Jinze and Ye, Rongguang and Ren, Dongwei},
  booktitle={AAAI},
  pages={12993--13000},
  year={2020}
}

@inproceedings{dai2017deformable,
  title={Deformable convolutional networks},
  author={Dai, Jifeng and Qi, Haozhi and Xiong, Yuwen and Li, Yi and Zhang, Guodong and Hu, Han and Wei, Yichen},
  booktitle={Proceedings of the IEEE international conference on computer vision},
  pages={764--773},
  year={2017}
}

@article{wu2020iou,
  title={IoU-aware single-stage object detector for accurate localization},
  author={Wu, Shengkai and Li, Xiaoping and Wang, Xinggang},
  journal={Image and Vision Computing},
  pages={103911},
  year={2020},
  publisher={Elsevier}
}

@inproceedings{he2016deep,
  title={Deep residual learning for image recognition},
  author={He, Kaiming and Zhang, Xiangyu and Ren, Shaoqing and Sun, Jian},
  booktitle={Proceedings of the IEEE conference on computer vision and pattern recognition},
  pages={770--778},
  year={2016}
}

@inproceedings{lin2017feature,
  title={Feature pyramid networks for object detection},
  author={Lin, Tsung-Yi and Doll{\'a}r, Piotr and Girshick, Ross and He, Kaiming and Hariharan, Bharath and Belongie, Serge},
  booktitle={Proceedings of the IEEE conference on computer vision and pattern recognition},
  pages={2117--2125},
  year={2017}
}

@inproceedings{girshick2015fast,
  title={Fast r-cnn},
  author={Girshick, Ross},
  booktitle={Proceedings of the IEEE international conference on computer vision},
  pages={1440--1448},
  year={2015}
}

@inproceedings{rezatofighi2019generalized,
  title={Generalized intersection over union: A metric and a loss for bounding box regression},
  author={Rezatofighi, Hamid and Tsoi, Nathan and Gwak, JunYoung and Sadeghian, Amir and Reid, Ian and Savarese, Silvio},
  booktitle={Proceedings of the IEEE Conference on Computer Vision and Pattern Recognition},
  pages={658--666},
  year={2019}
}

@inproceedings{lin2014microsoft,
  title={Microsoft coco: Common objects in context},
  author={Lin, Tsung-Yi and Maire, Michael and Belongie, Serge and Hays, James and Perona, Pietro and Ramanan, Deva and Doll{\'a}r, Piotr and Zitnick, C Lawrence},
  booktitle={European conference on computer vision},
  pages={740--755},
  year={2014},
  organization={Springer}
}

@inproceedings{zitnick2014edge,
  title={Edge boxes: Locating object proposals from edges},
  author={Zitnick, C Lawrence and Doll{\'a}r, Piotr},
  booktitle={European conference on computer vision},
  pages={391--405},
  year={2014},
  organization={Springer}
}

@article{zhang2019accurate,
  title={Accurate face detection for high performance},
  author={Zhang, Faen and Fan, Xinyu and Ai, Guo and Song, Jianfei and Qin, Yongqiang and Wu, Jiahong},
  journal={arXiv preprint arXiv:1905.01585},
  year={2019}
}

@inproceedings{szegedy2015going,
  title={Going deeper with convolutions},
  author={Szegedy, Christian and Liu, Wei and Jia, Yangqing and Sermanet, Pierre and Reed, Scott and Anguelov, Dragomir and Erhan, Dumitru and Vanhoucke, Vincent and Rabinovich, Andrew},
  booktitle={Proceedings of the IEEE conference on computer vision and pattern recognition},
  pages={1--9},
  year={2015}
}

@article{tian2018learning,
  title={Learning better features for face detection with feature fusion and segmentation supervision},
  author={Tian, Wanxin and Wang, Zixuan and Shen, Haifeng and Deng, Weihong and Meng, Yiping and Chen, Binghui and Zhang, Xiubao and Zhao, Yuan and Huang, Xiehe},
  journal={arXiv preprint arXiv:1811.08557},
  year={2018}
}

@article{zhang2019improved,
  title={Improved selective refinement network for face detection},
  author={Zhang, Shifeng and Zhu, Rui and Wang, Xiaobo and Shi, Hailin and Fu, Tianyu and Wang, Shuo and Mei, Tao and Li, Stan Z},
  journal={arXiv preprint arXiv:1901.06651},
  year={2019}
}

@article{zhang2019robust,
  title={Robust and high performance face detector},
  author={Zhang, Yundong and Xu, Xiang and Liu, Xiaotao},
  journal={arXiv preprint arXiv:1901.02350},
  year={2019}
}

@InProceedings{Liu_2019_CVPR,
author = {Liu, Wei and Liao, Shengcai and Ren, Weiqiang and Hu, Weidong and Yu, Yinan},
title = {High-Level Semantic Feature Detection: A New Perspective for Pedestrian Detection},
booktitle = {Proceedings of the IEEE/CVF Conference on Computer Vision and Pattern Recognition (CVPR)},
month = {June},
year = {2019}
}

@article{yoo2019extd,
  title={Extd: Extremely tiny face detector via iterative filter reuse},
  author={Yoo, YoungJoon and Han, Dongyoon and Yun, Sangdoo},
  journal={arXiv preprint arXiv:1906.06579},
  year={2019}
}

@inproceedings{chi2019selective,
  title={Selective refinement network for high performance face detection},
  author={Chi, Cheng and Zhang, Shifeng and Xing, Junliang and Lei, Zhen and Li, Stan Z and Zou, Xudong},
  booktitle={Proceedings of the AAAI Conference on Artificial Intelligence},
  volume={33},
  pages={8231--8238},
  year={2019}
}

@article{wang2017face,
  title={Face attention network: An effective face detector for the occluded faces},
  author={Wang, Jianfeng and Yuan, Ye and Yu, Gang},
  journal={arXiv preprint arXiv:1711.07246},
  year={2017}
}

@article{zhang2020feature,
  title={Feature agglomeration networks for single stage face detection},
  author={Zhang, Jialiang and Wu, Xiongwei and Hoi, Steven CH and Zhu, Jianke},
  journal={Neurocomputing},
  volume={380},
  pages={180--189},
  year={2020},
  publisher={Elsevier}
}

@article{zhang2019single,
  title={Single-shot scale-aware network for real-time face detection},
  author={Zhang, Shifeng and Wen, Longyin and Shi, Hailin and Lei, Zhen and Lyu, Siwei and Li, Stan Z},
  journal={International Journal of Computer Vision},
  volume={127},
  number={6-7},
  pages={537--559},
  year={2019},
  publisher={Springer}
}

@article{zhang2018face,
  title={Face detection using improved faster rcnn},
  author={Zhang, Changzheng and Xu, Xiang and Tu, Dandan},
  journal={arXiv preprint arXiv:1802.02142},
  year={2018}
}

@article{wang2017detecting,
  title={Detecting faces using region-based fully convolutional networks},
  author={Wang, Yitong and Ji, Xing and Zhou, Zheng and Wang, Hao and Li, Zhifeng},
  journal={arXiv preprint arXiv:1709.05256},
  year={2017}
}

@InProceedings{Zhu_2018_CVPR,
author = {Zhu, Chenchen and Tao, Ran and Luu, Khoa and Savvides, Marios},
title = {Seeing Small Faces From Robust Anchor's Perspective},
booktitle = {Proceedings of the IEEE Conference on Computer Vision and Pattern Recognition (CVPR)},
month = {June},
year = {2018}
}

@InProceedings{Najibi_2017_ICCV,
author = {Najibi, Mahyar and Samangouei, Pouya and Chellappa, Rama and Davis, Larry S.},
title = {SSH: Single Stage Headless Face Detector},
booktitle = {Proceedings of the IEEE International Conference on Computer Vision (ICCV)},
month = {Oct},
year = {2017}
}

@article{wang2017facercnn,
  title={Face r-cnn},
  author={Wang, Hao and Li, Zhifeng and Ji, Xing and Wang, Yitong},
  journal={arXiv preprint arXiv:1706.01061},
  year={2017}
}

@InProceedings{Hu_2017_CVPR,
author = {Hu, Peiyun and Ramanan, Deva},
title = {Finding Tiny Faces},
booktitle = {Proceedings of the IEEE Conference on Computer Vision and Pattern Recognition (CVPR)},
month = {July},
year = {2017}
}

@inproceedings{cai2016unified,
  title={A unified multi-scale deep convolutional neural network for fast object detection},
  author={Cai, Zhaowei and Fan, Quanfu and Feris, Rogerio S and Vasconcelos, Nuno},
  booktitle={European conference on computer vision},
  pages={354--370},
  year={2016},
  organization={Springer}
}

@article{yang2017face,
  title={Face detection through scale-friendly deep convolutional networks},
  author={Yang, Shuo and Xiong, Yuanjun and Loy, Chen Change and Tang, Xiaoou},
  journal={arXiv preprint arXiv:1706.02863},
  year={2017}
}

@incollection{zhu2017cms,
  title={Cms-rcnn: contextual multi-scale region-based cnn for unconstrained face detection},
  author={Zhu, Chenchen and Zheng, Yutong and Luu, Khoa and Savvides, Marios},
  booktitle={Deep learning for biometrics},
  pages={57--79},
  year={2017},
  publisher={Springer}
}

@article{zhang2016joint,
  title={Joint face detection and alignment using multitask cascaded convolutional networks},
  author={Zhang, Kaipeng and Zhang, Zhanpeng and Li, Zhifeng and Qiao, Yu},
  journal={IEEE Signal Processing Letters},
  volume={23},
  number={10},
  pages={1499--1503},
  year={2016},
  publisher={IEEE}
}

@inproceedings{ohn2016boost,
  title={To boost or not to boost? on the limits of boosted trees for object detection},
  author={Ohn-Bar, Eshed and Trivedi, Mohan M},
  booktitle={2016 23rd international conference on pattern recognition (ICPR)},
  pages={3350--3355},
  year={2016},
  organization={IEEE}
}

@inproceedings{yang2015facial,
  title={From facial parts responses to face detection: A deep learning approach},
  author={Yang, Shuo and Luo, Ping and Loy, Chen-Change and Tang, Xiaoou},
  booktitle={Proceedings of the IEEE international conference on computer vision},
  pages={3676--3684},
  year={2015}
}

@inproceedings{yang2014aggregate,
  title={Aggregate channel features for multi-view face detection},
  author={Yang, Bin and Yan, Junjie and Lei, Zhen and Li, Stan Z},
  booktitle={IEEE international joint conference on biometrics},
  pages={1--8},
  year={2014},
  organization={IEEE}
}

@inproceedings{felzenszwalb2008discriminatively,
  title={A discriminatively trained, multiscale, deformable part model},
  author={Felzenszwalb, Pedro and McAllester, David and Ramanan, Deva},
  booktitle={2008 IEEE conference on computer vision and pattern recognition},
  pages={1--8},
  year={2008},
  organization={IEEE}
}

@inproceedings{dalal2005histograms,
  title={Histograms of oriented gradients for human detection},
  author={Dalal, Navneet and Triggs, Bill},
  booktitle={2005 IEEE computer society conference on computer vision and pattern recognition (CVPR'05)},
  volume={1},
  pages={886--893},
  year={2005},
  organization={IEEE}
}

@article{lowe2004distinctive,
  title={Distinctive image features from scale-invariant keypoints},
  author={Lowe, David G},
  journal={International journal of computer vision},
  volume={60},
  number={2},
  pages={91--110},
  year={2004},
  publisher={Springer}
}

@article{krizhevsky2017imagenet,
  title={Imagenet classification with deep convolutional neural networks},
  author={Krizhevsky, Alex and Sutskever, Ilya and Hinton, Geoffrey E},
  journal={Communications of the ACM},
  volume={60},
  number={6},
  pages={84--90},
  year={2017},
  publisher={AcM New York, NY, USA}
}

@inproceedings{girshick2014rich,
  title={Rich feature hierarchies for accurate object detection and semantic segmentation},
  author={Girshick, Ross and Donahue, Jeff and Darrell, Trevor and Malik, Jitendra},
  booktitle={Proceedings of the IEEE conference on computer vision and pattern recognition},
  pages={580--587},
  year={2014}
}

@inproceedings{ren2015faster,
  title={Faster r-cnn: Towards real-time object detection with region proposal networks},
  author={Ren, Shaoqing and He, Kaiming and Girshick, Ross and Sun, Jian},
  booktitle={Advances in neural information processing systems},
  pages={91--99},
  year={2015}
}

@inproceedings{cai2018cascade,
  title={Cascade r-cnn: Delving into high quality object detection},
  author={Cai, Zhaowei and Vasconcelos, Nuno},
  booktitle={Proceedings of the IEEE conference on computer vision and pattern recognition},
  pages={6154--6162},
  year={2018}
}

@inproceedings{liu2016ssd,
  title={Ssd: Single shot multibox detector},
  author={Liu, Wei and Anguelov, Dragomir and Erhan, Dumitru and Szegedy, Christian and Reed, Scott and Fu, Cheng-Yang and Berg, Alexander C},
  booktitle={European conference on computer vision},
  pages={21--37},
  year={2016},
  organization={Springer}
}

@article{simonyan2014very,
  title={Very deep convolutional networks for large-scale image recognition},
  author={Simonyan, Karen and Zisserman, Andrew},
  journal={arXiv preprint arXiv:1409.1556},
  year={2014}
}

@inproceedings{he2017mask,
  title={Mask r-cnn},
  author={He, Kaiming and Gkioxari, Georgia and Doll{\'a}r, Piotr and Girshick, Ross},
  booktitle={Proceedings of the IEEE international conference on computer vision},
  pages={2961--2969},
  year={2017}
}

@inproceedings{viola2001robust,
  title={Robust real-time face detection},
  author={Viola, Paul and Jones, Michael},
  booktitle={null},
  pages={747},
  year={2001},
  organization={IEEE}
}

@inproceedings{redmon2016you,
  title={You only look once: Unified, real-time object detection},
  author={Redmon, Joseph and Divvala, Santosh and Girshick, Ross and Farhadi, Ali},
  booktitle={Proceedings of the IEEE conference on computer vision and pattern recognition},
  pages={779--788},
  year={2016}
}

@inproceedings{redmon2017yolo9000,
  title={YOLO9000: better, faster, stronger},
  author={Redmon, Joseph and Farhadi, Ali},
  booktitle={Proceedings of the IEEE conference on computer vision and pattern recognition},
  pages={7263--7271},
  year={2017}
}

@article{redmon2018yolov3,
  title={Yolov3: An incremental improvement},
  author={Redmon, Joseph and Farhadi, Ali},
  journal={arXiv preprint arXiv:1804.02767},
  year={2018}
}

@inproceedings{singh2018analysis,
  title={An analysis of scale invariance in object detection snip},
  author={Singh, Bharat and Davis, Larry S},
  booktitle={Proceedings of the IEEE conference on computer vision and pattern recognition},
  pages={3578--3587},
  year={2018}
}

@inproceedings{singh2018sniper,
  title={Sniper: Efficient multi-scale training},
  author={Singh, Bharat and Najibi, Mahyar and Davis, Larry S},
  booktitle={Advances in neural information processing systems},
  pages={9310--9320},
  year={2018}
}

\end{document}